\documentclass[11pt,a4paper]{article}
\usepackage[hyperref]{acl2020}
\usepackage{times}
\usepackage{latexsym}

\usepackage{xspace}
\usepackage{booktabs}
\usepackage{multicol}
\usepackage{CJKutf8}
\usepackage{graphicx}
\usepackage{multirow}

\usepackage{microtype}

\aclfinalcopy 

\newcommand{\ie}{i.e.,\xspace}
\newcommand{\eg}{e.g.,\xspace}
\newcommand{\eat}[1]{}
\newcommand{\paratitle}[1]{\noindent\textbf{#1}}

\newcommand{\baby}{\textsc{MatInf}\xspace}
\newcommand{\subbaby}[1]{\textsc{MatInf-#1}\xspace}
\newcommand{\mt}{\textsc{MTF-S2S}\xspace}

\def\aclpaperid{811} 

\title{\baby: A Jointly Labeled Large-Scale Dataset for Classification, Question Answering and Summarization}

 \author{Canwen Xu$^1$\thanks{\ \ \ The first two authors contribute equally to this paper.} , Jiaxin Pei$^2$$^*$, Hongtao Wu$^3$, Yiyu Liu$^3$, Chenliang Li$^3$\thanks{\ \ \ Chenliang Li is the corresponding author.}\\
 $^1$ School of Computer Science, Wuhan University, China \\
 $^2$ School of Information, University of Michigan, United States \\
 $^3$ School of Cyber Science and Engineering, Wuhan University, China \\
 $^{1,3}$ {\tt \{xucanwen,wuhongtao,liuyiyu,cllee\}@whu.edu.cn} \\
 $^2$ {\tt pedropei@umich.edu}
}

\date{}

\begin{document}
\maketitle
\begin{abstract}
Recently, large-scale datasets have vastly facilitated the development in nearly all domains of Natural Language Processing. However, there is currently no cross-task dataset in NLP, which hinders the development of multi-task learning. We propose \baby, the first jointly labeled large-scale dataset for classification, question answering and summarization. \baby contains 1.07 million question-answer pairs with human-labeled categories and user-generated question descriptions. Based on such rich information, \baby is applicable for three major NLP tasks, including classification, question answering, and summarization. We benchmark existing methods and a novel multi-task baseline over \baby to inspire further research. Our comprehensive comparison and experiments over \baby and other datasets demonstrate the merits held by \baby. \footnote{The implementation of \mt and information about obtaining access to the dataset can be found at \url{https://github.com/WHUIR/MATINF}.}
\end{abstract}

\section{Introduction}
In recent years, large-scale datasets (\eg ImageNet~\cite{imagenet} and SQuAD~\cite{squad}) have inspired remarkable progress in many areas like Computer Vision (CV) and Natural Language Processing (NLP). On the one hand, well-annotated data provide essential information for training supervised machine learning models. On the other hand, benchmarked datasets make it possible to evaluate and compare the capability of different methods on the same stage. 

Due to the high cost of data annotation, existing NLP datasets are usually labeled for only one particular task~(\eg SQuAD~\cite{squad} for question answering, CNN/DM~\cite{cnndm} for summarization and AGNews~\cite{agnewsNdbpediaNyahoo} for text classification). These single-task datasets hinder the development of learning common and task-invariant knowledge~\cite{adversial}. Although multi-task learning and transfer learning have delivered encouraging results, we still cannot determine whether the improvement is from the extension of input or supervision. Furthermore, task-specific data make the models tend to learn task-specific leakage features~\cite{leakage} rather than meaningful knowledge that could generalize to other tasks. However, as a key step to Artificial General Intelligence (AGI), knowledge acquisition requires the model to learn more general knowledge instead of overfitting on a specific task. Therefore, a large-scale and cross-task dataset is in huge demand for future NLP research. Nevertheless, to the best of our knowledge, none of the existing datasets could meet such demand.

In this paper, we propose \textbf{Mat}ernal and \textbf{Inf}ant Dataset (\baby), the first large-scale dataset covering three major NLP tasks: text classification, question answering and summarization. \baby consists of question answering data crawled from a large Chinese maternity and baby caring QA site. On this site, users can ask questions related to maternity and baby caring. When submitting a question, a detailed description is required to provide essential information and the asker also needs to assign a category for this question from a pre-defined topic list. Each user could submit an answer to a question post, and the asker will select the best answer out of all the candidates. To attract more attention, the askers are encouraged to set rewards using virtual coins when submitting the question and these coins will be given to the user who submitted the best answer selected by the asker. This rewarding mechanism could constantly ensure high-quality answers.

\baby supports three NLP tasks as follows.
\paratitle{Text Classification.} Given a question and its detailed description, the task is to select an appropriate category from the fine-grained category list. Different from previous news classification tasks whose category set is general topics like entertainment and sports, \subbaby{C} is a fine-grained classification under a single domain. That is, the distance between different categories is smaller, which provides a more challenging stage to test the continuously evolving state-of-the-art neural models.

\paratitle{Question Answering.} Given a question, the task is to produce an answer in natural language. This task is slightly different from previous Machine Reading Comprehension (MRC) since the document which contains the correct answer is not directly provided. Therefore, how to collect the domain knowledge from massive QA data becomes extremely important. 

\paratitle{Summarization.} Given a question description, the task is to produce the corresponding question. Previous summarization datasets are all constructed with news or academic articles. The limited text genres covered in these datasets hinder the thorough evaluation of summarization models. Also, the noisy nature of \baby encourages more robust models. \baby can be considered as the first social media summarization dataset.

\baby holds the following merits: (1) \textbf{Large}. \baby includes 1.07M unique QA pairs, making it an ideal playground for the new advancements of deeper and larger models (\eg Pretrained Language Models). (2) \textbf{Multi-task applicable}. \baby is the first dataset that simultaneously contains ground truths for three major NLP tasks, which could facilitate new multi-task learning methods for these tasks. Here, to set a baseline and inspire future research, we present \textbf{M}ulti-\textbf{t}ask \textbf{F}ield-shared \textbf{S}equence \textbf{to} \textbf{S}equence (\mt), a straightforward yet effective model, which achieves better performance on all three tasks compared to its single-task counterparts.

\section{Related Work}
\subsection{Topic Classification}
\eat{
As a fundamental problem in NLP, text classification has been deeply researched in the past decade with consistently emerged novel methods. With the prevalence of Deep Neural Networks, the task is dominated by neural network architectures (\eg Convolutional Neural Networks (CNN)~\cite{dcnn,rcnn,agnewsNdbpediaNyahoo,dpcnn,vdcnn}, Recurrent Neural Networks (RNN)~\cite{icml16:johnson}, Transformer~\cite{transformer}). Also, Pretrained Language Models (PLMs) show dominant performance on this task~\cite{ulmfit,bert,xlnet}. 
}

Topic classification is one of the most fundamental tasks in NLP. As a deeply explored task, many datasets have been used in previous research both in English (AGNews, DBPedia, Yahoo Answer~\cite{agnewsNdbpediaNyahoo}, TREC~\cite{trec}) and Chinese (THUCNews~\cite{thucnews}, SogouCS~\cite{sogou}, Fudan Corpus, iFeng and ChinaNews~\cite{ifengNchinanews}). These datasets were useful and indispensable in the past decades to test the performance of different kinds of classifiers. 

However, as most of them are formal text and the target categories are general topics, even simply leveraging n-gram features could achieve acceptable results. Plus, some of them are small in scale. Nowadays, with the prevalence of neural models and pretraining techniques, recent algorithms~\cite{superchar,glyce} are approaching the ceiling of these datasets with accuracy scores up to $98\%$. Different from any of the existing datasets, \baby is more challenging, providing a new stage to test the performance of future algorithms.

\subsection{Question Answering}
Following the definition in \cite{jurafsky2008speech}, Question Answering (QA) can be generally divided into Information Retrieval (IR) based Question Answering and Knowledge-based Question Answering. For IR-based Question Answering, the answer is often a span in the retrieved document. As for Knowledge-based Question Answering, a human-constructed knowledge base is provided for querying and the answer is in the form of a query result. Recently, Open Domain QA~\cite{chen2017reading} has been recognized as a new genre where a natural language response instead of text spans is returned as an answer.

Currently, several datasets are available for Chinese Question Answering. NLPCC Shared Task~\cite{nlpcc17:duan} provided two datasets for IR-based and Knowledge-based QA, respectively. DuReader~\cite{dureader} is an Open Domain dataset derived from user search logs and provided with human-picked documents as evidence. \citet{gaokao} provided a QA dataset in the domain of Chinese College Entrance Test history exam questions, with documents from standard history textbooks. Different from these datasets, instead of providing pre-defined documents as evidence, \subbaby{QA} only contains sufficient QA pairs in the training set. In this way, there are various approaches to exploit these questions as evidence. Thus, \subbaby{QA} encourages innovations in retrieval, generation and hybrid question answering methods.

\subsection{Summarization}
Summarization datasets can be roughly categorized into extractive and abstractive datasets, which respectively favor abstractive and extractive methods. 
Extractive datasets are composed of long documents and summaries. Since the summary is long, extracted sentences and spans from the document could compose a good summary. Newsroom~\cite{newsroom}, ArXiv and PubMed~\cite{arxiv} and CNN / Daily Mail dataset~\cite{cnndm} are commonly used extractive datasets.

Abstractive datasets often contain short documents and summaries, which encourages a thorough understanding of the document and style transfer between a document and its corresponding summary. Gigaword~\cite{gigaword} and Xsum~\cite{xsum} fall into this category. Also, the abstractive dataset LCSTS~\cite{lcsts}, crawled from verified short news feeds of major newspapers and televisions, is the only public dataset for Chinese text summarization to date.

However, all of these existing datasets are composed of either news or academic articles. The narrow sources of these datasets bring two main drawbacks. First, due to the nature of news reporting and academic writing, the summary-eligible contents do not distribute uniformly~\cite{bigpatent}. Second, models evaluated on these noiseless formal-text datasets are not robust enough for real-world applications. To address these problems, we propose \subbaby{Summ}, a new abstractive Chinese summarization dataset.

\begin{figure}[t]
\centering
\includegraphics[width=\columnwidth]{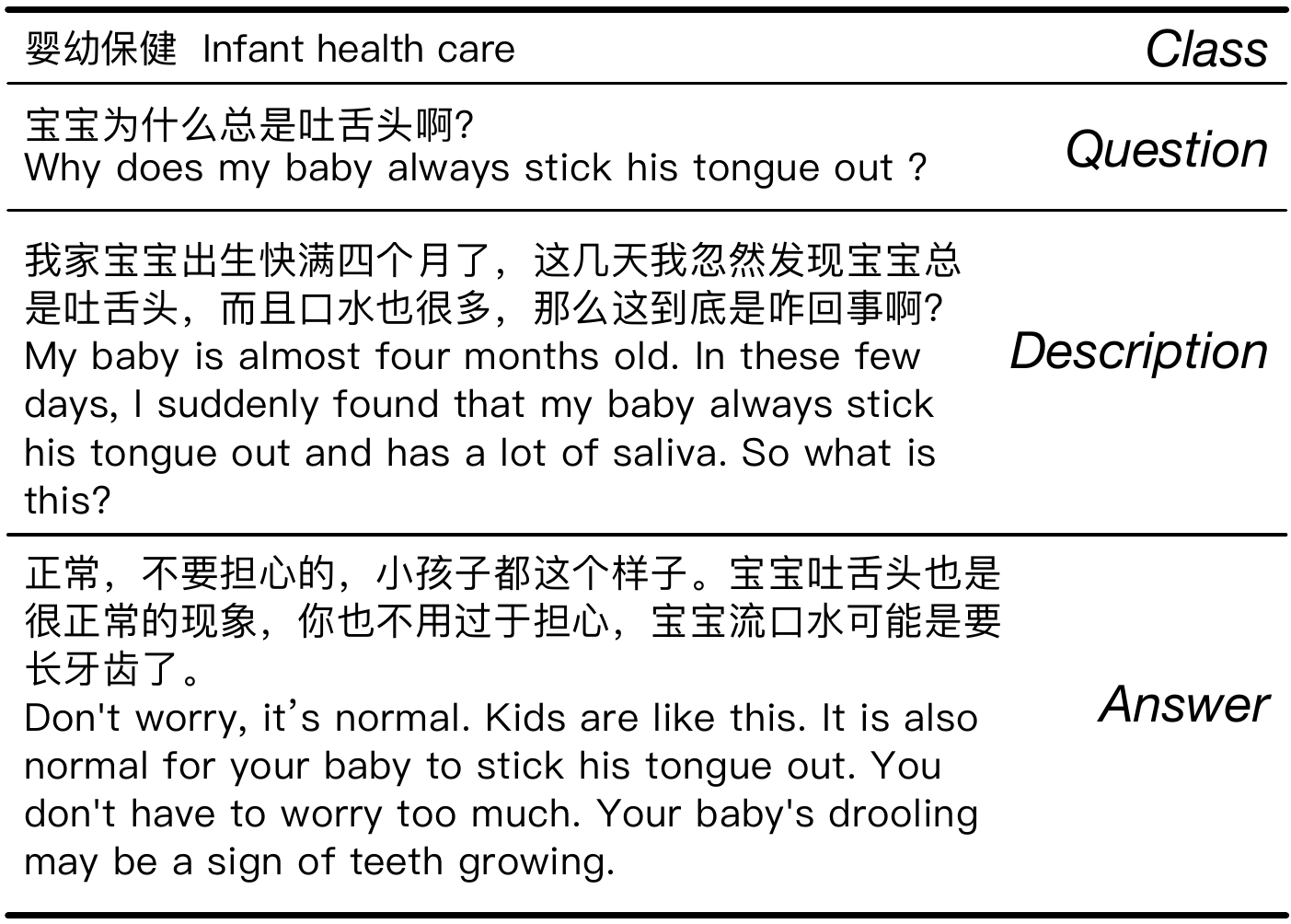}
\caption{An example entry from \baby.}
\label{example}
\end{figure}

\begin{table}[t]
\centering
\resizebox{\columnwidth}{!}{
\smallskip\begin{tabular}{lccc|c}
\toprule
& Question & Description & Answer & Max Len. \\
\midrule
\#~Char & 14.72 & 64.17 & 66.91 & 256\\
\#~Word & 9.03 & 41.70 & 42.32 & -\\
\bottomrule
\end{tabular}
}
\caption{\label{babystat}Average character and word numbers of question, description and answer in \baby. We ensure that every field of each entry has at most $256$ characters.}
\end{table}

\section{\baby Dataset}
We present \textbf{Mat}ernal and \textbf{Inf}ant (\baby) Dataset, a large-scale dataset jointly labeled for classification, question answering and summarization in the domain of maternity and baby caring in Chinese.  An entry in the dataset includes four fields: \textit{question~(Q)}, \textit{description~(D)}, \textit{class~(C)} and \textit{answer~(A)}. An example is shown in Figure \ref{example}, and the average character and word numbers of each field are reported in Table \ref{babystat}. 

We collect nearly two million question-answer pairs with fine-grained human-labeled classes from a large Chinese maternity and baby caring QA site. We conduct both automatic and manual data cleansing and remove: (1) classes with insufficient samples; (2) entries in which the length of the description filed is less than the length of the question field; (3) data with any field longer than $256$ characters; (4) human-spotted ill-formed data. After the data cleansing, we construct \baby with the remaining $1.07$ million entries.

We first randomly split the whole data into training, validation and test sets with a proportion of 7:1:2. Then, we use the splits for summarization and QA. For classification, we further divide the data into two sub-tasks according to different classification standards within each split.

\subsection{\subbaby{C}: Fine-grained Text Classification}
In \baby, the class labels are first selected by the users when submitting a question. Then, if the question is not in the right class, the forum administrators would manually re-categorize the question to the correct class. In our data, there are two parallel standards for classifying a question: \textit{topic class} and \textit{age of the baby}. We use these two standards to construct our two subsets. Thus, we define two tasks: (1) classifying a question to different age groups; (2) classifying a question into a fine-grained topic. We list the classes of the two tasks in Table \ref{classnames}. Note that there is no data overlap between the two subsets. Formally, we define the task as predicting the class of a QA pair with its question and description fields (\ie $Q, D \rightarrow C$). Different from previous datasets, our task is a fine-grained classification (\ie to classify documents in a domain) rather than classifying general topics (\eg politics, sports, entertainments), which means the semantic difference between classes is prominently smaller. It requires meticulous exploitation of semantics instead of recognizing unique n-gram features for each class. We provide statistical comparison of \subbaby{C} with other datasets in Table \ref{classtat}.

\begin{CJK*}{UTF8}{gbsn}
\begin{table}[t]
\centering
\resizebox{\columnwidth}{!}{
\smallskip\begin{tabular}{l|l||l|l}
\toprule
\multicolumn{2}{c||}{\subbaby{C-Topic}} &
\multicolumn{2}{c}{\subbaby{C-Age}} \\
\multicolumn{2}{c||}{18 classes} &
\multicolumn{2}{c}{3 classes} \\
\midrule
产褥期保健 & postpartum health care & 0-1岁 & 0-1 yr old\\
儿童过敏 & child allergy & 1-2岁 & 1-2 yrs old\\
动作发育 & motion development & 2-3岁 & 2-3 yrs old\\
婴幼保健 & infant health care & & \\
婴幼心理 & infant psychology & & \\
婴幼早教 & early education & & \\
婴幼期喂养 & infant feeding & & \\
婴幼营养 & infant nutrition & & \\
孕期保健 & pregnancy care & & \\
家庭教育 & family education & & \\
幼儿园 & kindergarten & & \\
未准父母 & pregnancy preparation & & \\
流产和不孕 & infertility problem & & \\
疫苗接种 & vaccination & & \\
皮肤护理 & skin care & & \\
宝宝上火 & infant ulcer & & \\
腹泻 & diarrhea & & \\
婴幼常见病 & other infant common diseases & & \\

\bottomrule
\end{tabular}
}
\caption{\label{classnames}Class names of two subsets and their English translations.}
\end{table}
\end{CJK*}

\begin{table}[t]
\centering
\resizebox{\columnwidth}{!}{
\smallskip\begin{tabular}{lllrr}
\toprule
Dataset & Lang. & Domain & \#~Doc & \#~Class\\
\midrule
AG News~\shortcite{agnewsNdbpediaNyahoo} & EN & News & 128K & 4 \\
DBPedia~\shortcite{agnewsNdbpediaNyahoo} & EN & Wiki & 630K & 14 \\
TREC-6~\shortcite{trec} & EN & Open & 6K & 6 \\
TREC-50 $^\dagger$~\shortcite{trec} & EN & Open & 6K & 50 \\
Yahoo Answer~\shortcite{agnewsNdbpediaNyahoo} & EN & Open & 1.46M & 10 \\
\midrule
THUCNews~\shortcite{thucnews} & ZH & News & 740K & 14 \\
SogouCS~\shortcite{sogoucs} & ZH & News & 577K & 5 \\
Fudan Corpus~\shortcite{cw2vec} & ZH & News & 10K & 20 \\
iFeng~\shortcite{ifengNchinanews} & ZH & News & 850K & 5 \\
ChinaNews~\shortcite{ifengNchinanews} & ZH & News & 1.51M & 7 \\
\midrule
\subbaby{C-Age} $^\dagger$ & ZH & Health & 192K & 3 \\
\subbaby{C-Topic} $^\dagger$ & ZH & Health & 876K & 18 \\

\bottomrule
\end{tabular}
}
\caption{\label{classtat}Comparison of classification datasets. \dag: Fine-grained datasets.}
\end{table}

\begin{table*}[t]
\centering
\resizebox{\textwidth}{!}{
\smallskip\begin{tabular}{llrrlll}
\toprule
Dataset & Lang. &\#~Q/A Pair &\#~Docs & Source of Query & Source of Docs & Answer Type\\
\midrule
CNN~/~DM~\shortcite{cnndm} & EN & 1.4M & 300K & Synthetic cloze & News & Fill in entity \\
HLF-RC~\shortcite{hlfrc} & ZH & 100K & 28K & Synthetic cloze & Fairy~/~News & Fill in word \\
CBT~\shortcite{cbt} & EN & 688K & 108 & Synthetic cloze & Children's books & Multi-choices \\
NewsQA~\shortcite{newsqa} & EN & 100K & 10K & Crowdsourced & CNN & Span of words \\
SQuAD~\shortcite{squad} & EN & 100K & 536 & Crowdsourced & Wiki & Span of words \\
SearchQA~\shortcite{searchqa} & EN & 140K & 6.9M & QA site & Web & Span of words \\
SQuAD 2.0~\shortcite{squad} & EN & 150K & 505 & Crowdsourced & Wiki & Span of words \\
NLPCC DBQA~\shortcite{nlpcc17:duan} & ZH & 15K & 15K & Crowdsourced & Wiki & Binary matching \\
MS-MARCO~\shortcite{msmarco} & EN & 100K & 200K & User logs & Web & Natural language response \\
DuReader~\shortcite{dureader} & ZH & 200K & 1M & User logs & Web/QA site & Natural language response \\

\midrule
\subbaby{QA} & ZH & 1.07M & - & QA Site & - & Natural language response \\
\bottomrule
\end{tabular}
}
\caption{\label{qastat}Comparison of question answering datasets. Some statistics are reused from \cite{dureader}.}
\end{table*}

\begin{table}[t]
\centering
\resizebox{\columnwidth}{!}{
\smallskip\begin{tabular}{lllrrr}
\toprule
Dataset & Lang. & Domain & \#~Doc & \multicolumn{2}{c}{\# Token}\\
& & & & Doc. & Sum.\\
\midrule
CNN~/~DM~\shortcite{cnndm} & EN & News & 312K & 781 & 56 \\
NYT~\shortcite{gigaword} & EN & News & 655K & 796 & 45 \\
NewsRoom~\shortcite{newsroom} & EN & News & 1.21M & 751 & 30 \\
BigPatent~\shortcite{bigpatent} & EN & Academic & 1.34M & 3573 & 117 \\
arXiv~\shortcite{arxiv} & EN & Academic & 216K & 6914 & 293\\
PubMed~\shortcite{arxiv} & EN & Academic & 133K & 3224 & 214\\
Gigawords~\shortcite{gigaword} & EN & News & 4.02M & 31 & 8 \\
LCSTS~\shortcite{lcsts} & ZH & News & 2.40M & 104 & 17 \\
XSum~\shortcite{xsum} & EN & News & 227K & 431 & 23 \\
\midrule
\subbaby{Summ} & ZH & Health & 1.07M & 42 & 9 \\

\bottomrule
\end{tabular}
}
\caption{\label{sumstat}Comparison of summarization datasets. ``\#Token'' indicates the average token numbers of a document and a summary for each dataset.}
\end{table}

\subsection{\subbaby{QA}: Health-Domain Question Answering}
Typically, to return an answer for a specific question, the model needs to retrieve from a pre-defined document set or query a manually-constructed knowledge base. MS-MARCO~\cite{msmarco} utilizes a search engine to pre-filter $10$ documents from the Internet and uses them as the document set. However, searching itself is a challenging task that significantly affects the final performance. On the other hand, in a real-world scenario, it is impossible to define a document set covering all knowledge needed to answer a user question. Thus, we provide the training set of \subbaby{QA} as the possible document source and encourage all kinds of methods including retrieval, generation and hybrid models.

Formally, the task is defined as replying a question with natural text (\ie $Q \rightarrow A$). The large scale of our dataset ensures that a model is able to generalize and learn enough knowledge to answer a user question. Note that we do not use description when defining this task since we observe a negative effect on the generalization in our experiment. Shown in Table \ref{qastat}, we list statistics of \subbaby{QA} and other commonly-used datasets.

\subsection{\subbaby{Summ}: Summarization in Professional Domain}
All current datasets for summarization to date are in the domain of news and academic articles. However, as a custom of the report and academic writing, in extractive datasets, the summary-eligible contents often appear at the beginning or the end of an article, preventing the summarization model from a full understanding and resulting in impractically high performance in evaluation. On the other hand, current abstractive datasets are all formal news datasets, which are in lack of diversity. Models trained on such a single-source dataset is not robust enough to handle real-world complexity.

In \subbaby{Summ}, question description can be seen as an extended and specific version of the question itself, containing more detailed background information with respect to the question. Besides, the question itself is often a well-formed interrogative sentence rather than extracted phrases. Our task is to generate the question from the corresponding description (\ie $D \rightarrow Q$). Note that our task itself can support many meaningful real-world applications, \eg generating an informative title for user-generated content (UGC). Also, there is only one public dataset for summarization in Chinese to date. Our dataset can be used to verify the effectiveness of existing models and eliminate the overfitting bias caused by evaluation on merely one dataset. We compare \subbaby{Summ} with other datasets in Table \ref{sumstat}.

\section{Multi-task Learning}

\begin{figure}[t]
\centering
\includegraphics[width=\columnwidth]{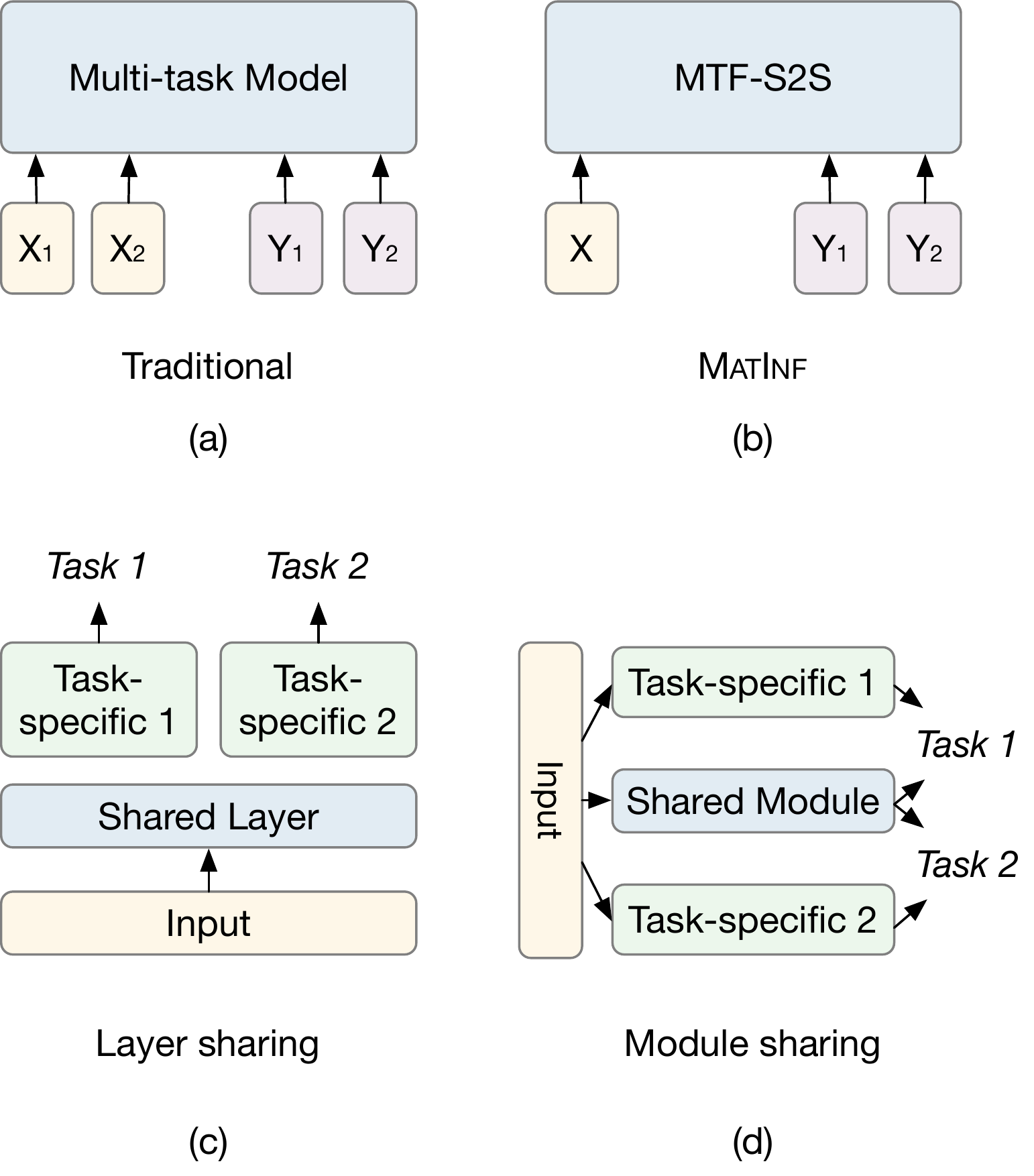}
\caption{The difference between \mt and traditional multi-task learning.}
\label{fig:multitask}
\end{figure}

\begin{figure}[t]
\centering
\includegraphics[width=\columnwidth]{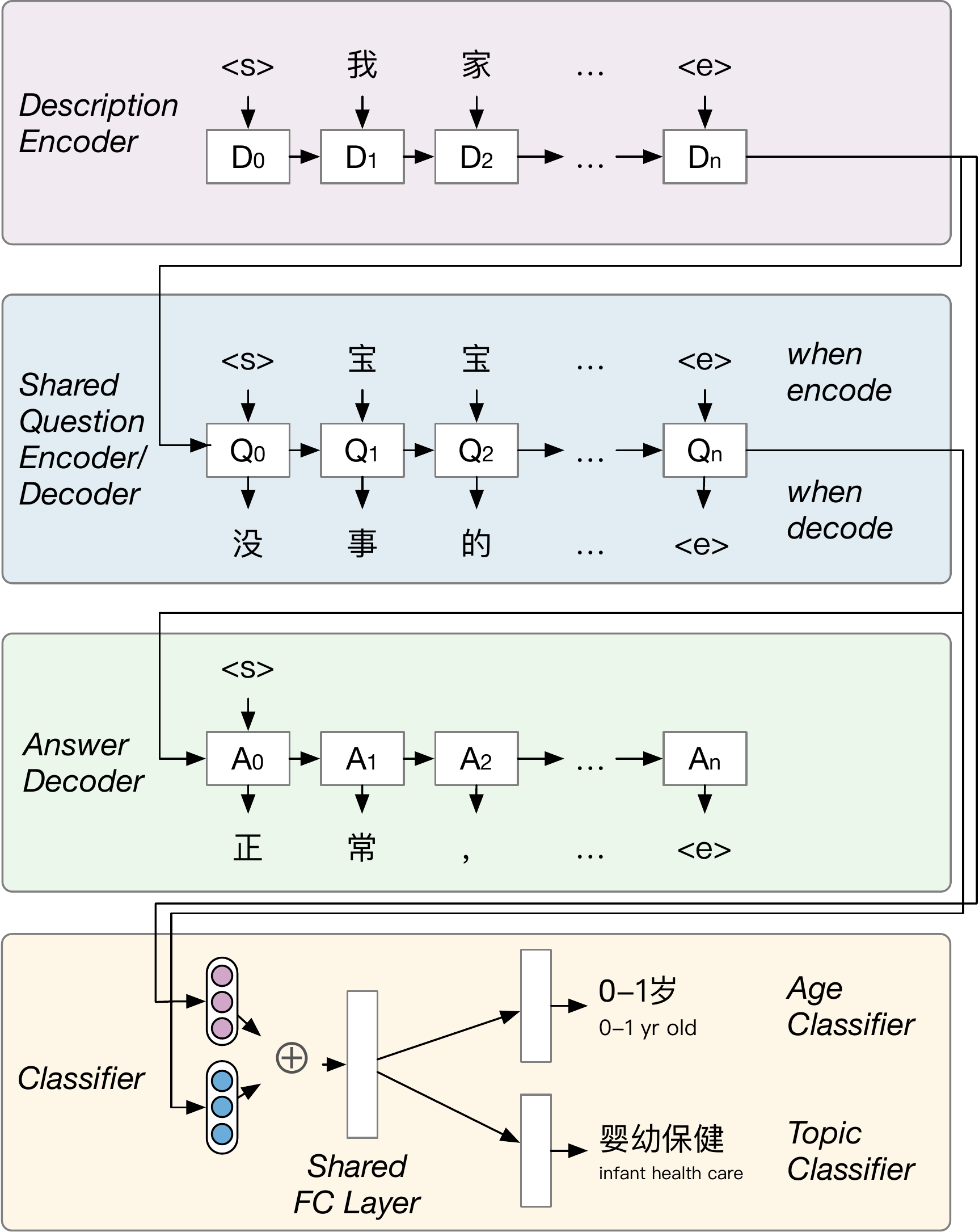}
\caption{The architecture of \mt. Note that a common attention mechanism~\cite{s2satt} is applied when decoding question and answer (in the blue and green boxes), but we do not illustrate it in this figure for clarity.}
\label{fig:mtf}
\end{figure}

Recently, many attempts have been made on multi-task learning in NLP~\cite{naacl15:liu,iclr16:luong,acl18:guo,decathlon,naacl19:xu,aaai19:ruder,mtdnn,gpt2,unilm,guoday,t5,xiangnan} and several benchmarks are available for multi-task evaluation~\cite{superglue,glue}.
Though recent studies show that multi-task learning is effective, there is still one more question to answer. That is, when training models on multiple tasks, multiple datasets are used by default. As illustrated in Figure \ref{fig:multitask}(a), it adds both new input (\ie text, denoted as $X$) and new supervision (\ie ground truths, denoted as $Y$). Due to the different processes of data collection, $X$ in different datasets have different sources and properties. Recent progress on Language Modeling~\cite{gpt2,bert,xlnet,t5} has proved that corpora ($X$) from different sources can make the model more robust and significantly improve the performance. To this end, it is not easy to determine whether the success of a multi-task model should be mainly attributed to the addition of $X$ or $Y$. However, as depicted in Figure \ref{fig:multitask}(b), in \baby, our jointly labeled fashion can guarantee that $X$ remains the same as in a single task and only $Y$ is added. Thus, \baby provides a fair and ideal stage for exploring multi-task learning, especially auxiliary and multi-task supervision under a single dataset.

To set a baseline and also inspire future research, we design a multi-task learning network, named \textbf{M}ulti-\textbf{t}ask \textbf{F}ield-shared \textbf{S}equence \textbf{to} \textbf{S}equence (\mt). We illustrate the architecture of \mt in Figure \ref{fig:mtf}. For generation tasks, we combine the summarization ($D \rightarrow Q$) and QA ($Q \rightarrow A$) to be the form of $D \rightarrow Q \rightarrow A$, with a shared Long Short-Term Memory (LSTM) for decoding questions in summarization task and encoding questions for both QA and classification tasks. Previous studies often share layers among tasks to regularize the representation learning, as illustrated in Figure \ref{fig:multitask}(c). Different from that, \mt shares on both module level (\ie field encoder/decoder, as shown in Figure \ref{fig:multitask}(d)) and layer level.
An attention mechanism is applied when decoding for summarization and QA. Also, we concatenate the encoded representations of description and question, and feed it to a shared fully connected layer and then specialized fully connected layers for age classification and topic classification, respectively.

When training, since the sizes of datasets for different tasks are not equal, we first determine the batch size for different tasks to make sure that the training progress for each task is approximately synchronized by:
\begin{equation}
	\forall a, b \in T, bs_a / bs_b = n_a / n_b
\end{equation}
where $T$ includes four tasks: summarization, QA, and two classification tasks. $bs_*$ is the batch size of each task, and $n_*$ is the sample numbers in each dataset for the task.
If one task is iterated to the last data batch, it will start over from the first batch.
For each iteration, we successively calculate the losses by Cross Entropy for each task in one batch.
Then, we train the model to minimize the total loss:
\begin{equation}
    \mathcal{L} = \sum_{t_i \in T}\lambda_i \mathcal{L}_i
\end{equation}
where $\lambda_*$ is the manually set weight for each task.
We stop the co-training after one epoch, then fine-tune the model to obtain the peak performance for each task, separately. 

\section{Experiments}
In this section, we benchmark a few baselines and \mt on the three tasks of \baby. We run each experiment with three different random seeds and report the average result of the three runs.

\subsection{Experimental Settings}
\paratitle{\mt.} For \mt, we set all $\lambda_i = 0.25$ and use an Adam~\cite{adam} optimizer to co-train the model for one epoch with batch sizes of $64$, $64$, $12$ and $52$ for $bs_{Summ}$, $bs_{QA}$, $bs_{CTopic}$, and $bs_{CAge}$ respectively with a learning rate of $0.001$. Then we fine-tune the model for each task with a learning rate of $5 \times 10^{-5}$. We report both the performance after co-training and after fine-tuning. The hidden size of all LSTM encoders/decoders and attentions is $200$. For all tasks, we separately train \mt on each task only to provide a single-task baseline. Both \mt and Seq2Seq baselines are character-based and their embeddings are initialized with Tencent AI Lab Embedding~\cite{tencentemb}. For both \mt and Seq2Seq baselines, we use Beam Search~\cite{beamsearch} when decoding.
\\ \\
\paratitle{Classification.} For classification, we conduct experiments with a statistical learning baseline, several deep neural networks and pretrained large-scale language models. For the statistical baselines, we extract character-based unigram and bigram features and use a logistic classifier to predict the classes.
For neural networks, we choose fastText~\cite{fasttext}, Text CNN~\cite{textcnn}, DCNN~\cite{dcnn}, RCNN~\cite{rcnn} and DPCNN~\cite{dpcnn}. As a classical step in Chinese text classification, we segment the sentences into words with Jieba\footnote{\url{https://github.com/fxsjy/jieba}. We use Jieba v0.39 throughout this paper.}, a commonly used out-of-the-box word segmentation toolkit. We then initialize the word embedding with pretrained Tencent AI Lab Embedding~\cite{tencentemb} except for fastText, which has its own algorithm to construct word embeddings. We minimize the Cross-Entropy with Adam~\cite{adam} optimizer with a learning rate of $0.001$ and apply early stopping.
For language models, we fine-tune BERT~\cite{bert} and ERNIE~\cite{ernie} that both have released official pretrained Chinese models. We set the learning rate for fine-tuning to $5 \times 10^{-5}$ and apply early stopping. We also compress the fine-tuned 12-layer BERT model with BERT-of-Theseus~\cite{theseus} and obtain the performance of a 6-layer model.

\paratitle{Question Answering.} For retrieval-based QA, following MS-MARCO~\cite{msmarco}, we calculate the average best scores between each answer in the test set and all answers in the training set within the same class, to determine the oracle retrieval performance. Then, we construct our retrieval-based baseline by fine-tuning BERT-Base~\cite{bert} for question matching on an external dataset, LCQMC~\cite{lcqmc}. Then we use the trained model to score the match between each question in the test set and all questions in the training set with the same class and return the answer of the top 1 matched question.
For generation-based baselines, we use character-based Seq2Seq~\cite{s2s} and Seq2Seq with Attention~\cite{s2satt}, since character-based method has a prominently better performance for Chinese text generation~\cite{lcsts,wordseg}. The metric for evaluation are ROUGE scores~\cite{rouge} calculated on the character level. 

\paratitle{Summarization.}
We categorize the baselines into two fashions: extractive methods (\ie extracting sentences or phrases from the text) and abstractive methods (\ie generating summaries according to the text).
For extractive methods, we choose two widely used classical methods, TextRank~\cite{textrank} and LexRank~\cite{lexrank}.
For abstractive methods, we use WEAN~\cite{wean} and Global Encoding~\cite{globalencoding} along with Seq2Seq~\cite{s2s,s2satt} as the baselines. We also add BertAbs~\cite{bertabs}, a BERT-based summarization model, to reflect the recent progress on this task. We use the officially released Chinese BERT-Base as the backbone. We use ROUGE scores~\cite{rouge} to evaluate the quality of generated summaries.

\begin{table}[t]
\centering
\resizebox{\columnwidth}{!}{
\smallskip\begin{tabular}{lcc}
\toprule
Method & AGE & TOPIC \\
\midrule
TF-IDF + LR$^\dagger$ & 76.88 & 40.25\\
\midrule
Text CNN~\cite{textcnn} & 90.95 & 64.41 \\
DCNN~\cite{dcnn}& 90.96 & 64.60 \\
RCNN~\cite{rcnn} & 90.81 & 63.56 \\
fastText~\cite{fasttext} & 87.76 & 61.81 \\
DPCNN~\cite{dpcnn} & \textbf{91.02} & 65.92 \\
\midrule
BERT$_{\rm{Base}}^\dagger$~\cite{bert} & 90.33 & \textbf{66.95} \\
BERT-of-Theseus$^\dagger$~\cite{theseus} & 90.25 & 66.72 \\
ERNIE~\cite{ernie} & 90.42 & 66.66 \\
\midrule
\mt~(single task)$^\dagger$ & 90.15 & 63.40 \\
\mt$^\dagger$ & 90.29 & 63.59 \\
\bottomrule
\end{tabular}
}
\caption{Experimental results of baseline methods on \subbaby{C} in terms of accuracy. \dag: Character-based models.}
\label{clasexp}
\end{table}

\begin{table}[t]
\centering
\resizebox{\columnwidth}{!}{
\smallskip\begin{tabular}{l|ccc}
\toprule
\multirow{2}{*}{Method} & \multicolumn{3}{c}{\subbaby{QA}} \\
& R-1 & R-2 & R-L \\
\midrule
Best Passage (upper bound) & 58.32 & 36.42 & 49.00  \\
\midrule
BERT Matching~\shortcite{bert} & 18.66 & 3.28 & 10.78 \\ 
\midrule
Seq2Seq~\shortcite{s2s} & 16.62 & 4.53 & 10.37 \\
Seq2Seq + Att~\shortcite{s2satt} & 19.62 & 5.87 & 13.34 \\
\midrule
\mt~(single task) & 20.28 & 5.94 & 13.52 \\
\mt & \textbf{21.66} & \textbf{6.58} & \textbf{14.26} \\

\bottomrule
\end{tabular}
}
\caption{Experimental results of baseline methods on \subbaby{QA}.}
\label{qaexp}
\end{table}

\begin{table*}[t]
\small
\centering
\smallskip\begin{tabular}{lccccccccc}
\toprule
& \multicolumn{3}{c}{CNN/DM} & \multicolumn{3}{c}{LCSTS} & \multicolumn{3}{c}{\subbaby{Summ}}\\
Method & R-1 & R-2 & R-L & R-1 & R-2 & R-L & R-1 & R-2 & R-L \\
\midrule
TextRank~\cite{textrank} & 37.72 & 15.59 & 33.81 & 24.38 & 11.97 & 16.76 & 35.53 & 25.78 & 36.84 \\
LexRank~\cite{lexrank} & 33.98 & 11.79 & 30.17 & 22.15 & 10.14 & 14.65 & 33.08 & 23.31 & 34.96 \\
\midrule
Seq2Seq~\cite{s2s} & - & - & - & - & - & - & 23.05 & 11.44 & 19.55 \\
Seq2Seq + Att~\cite{s2satt} & 31.33 & 11.81 & 28.83 & 33.80 & 23.10 & 32.50 & 43.05 & 28.03 & 38.58 \\
WEAN~\cite{wean} & - & - & - & 37.80 & 25.60 & 35.20 & 34.63 & 22.56 & 28.92 \\
Global Encoding~\cite{globalencoding} & - & - & - & \textbf{39.40} & \textbf{26.90} & \textbf{36.50} & 49.28 & 34.14 & 47.64 \\
BertAbs~\cite{bertabs} & \textbf{40.21} & \textbf{17.76} & \textbf{37.09} & - & - & - & \textbf{57.31} & \textbf{44.05} & \textbf{55.93} \\
\midrule
\mt~(single task) & 31.36 & 11.80 & 28.88 & 33.75 & 23.20 & 32.51 & 43.02 & 28.05 & 38.55 \\
\mt & - & - & - & - & - & - & 48.59 & 35.69 & 43.28 \\
\bottomrule
\end{tabular}
\caption{Experimental results of baseline methods on CNN~/~DM~\cite{cnndm}, LCSTS~\cite{lcsts}, and \subbaby{Summ}.}
\label{sumexp}
\end{table*}

\subsection{Results and Analysis}

\paratitle{Classification.}
We show the experimental results of two classification sub-tasks in Table \ref{clasexp}. On the tougher \subbaby{C-Topic}, language models prominently outperform other baselines. Among non-LM neural networks, DPCNN~\cite{dpcnn}, which has the deepest architecture and the most parameters, outperforms other baselines with a considerable margin. On \subbaby{C-Age}, which is a smaller dataset with fewer classes, DPCNN outperforms all other baselines including language models with an accuracy of $91.02$. To analyze, this task has fewer training samples, which is in favor of a model with moderate parameter numbers instead of huge parameter numbers as in language models. Also, the task is relatively easier due to the class number, which makes the advantage of language models more trivial. For the multi-task baseline, \mt shows a satisfying performance on both \subbaby{C-Age} and \subbaby{C-Topic}, outperforming the same model which is only trained on the single task by $0.14$ and $0.19$ in terms of accuracy. Notably, BERT-of-Theseus~\cite{theseus} has a satisfying performance compressing the fine-tuned BERT to smaller models.

\paratitle{Question Answering.}
The experimental results are shown in Table \ref{qaexp}. The high scores of Best Passage (maximum possible performance) indicate that using training data as a document set is completely feasible. Seq2Seq with Attention outperforms the retrieval-based baseline by a margin of $2.56$ in terms of ROUGE-L. It suggests that a generation-based neural network can effectively learn from multiple relevant samples and generalize. Besides, since we do the matching between each question and every entry within the same class in the training set, the inference of BERT Matching takes quite a long time. Similar to MS-MARCO~\cite{msmarco}, it is possible to use a search engine (\eg Elastic Search) to pre-filter the documents and reduce the computational cost. Meanwhile, \mt is effective on QA task and outperforms its single-task version by $0.74$ on ROUGE-L.

\paratitle{Summarization.} 
We further conduct performance comparison for summarization across three datasets, CNN/DM~\cite{cnndm}, LCSTS~\cite{lcsts}, and our \subbaby{Summ} in Table \ref{sumexp}. By comparing the performance of two basic baselines, TextRank~\cite{textrank} and Seq2Seq+Att~\cite{s2satt}, we can see an obvious difference in performance between extractive and abstractive methods on datasets of different genres. 
BertAbs~\cite{bertabs}, the powerful BERT-based model, significantly outperforms all other baselines on \subbaby{Summ} thanks to its exploitation of pretraining and the capacity of a BERT model. For \mt, it outperforms the single-task counterpart by $4.73$ on ROUGE-L.

\section{Discussion}
Since \baby is a web-crawled dataset, it would be inevitable to be noisier than a dataset annotated by hired annotators though we have made every effort to clean the data. On the bright side, it can encourage more robust models and facilitate real-world applications. For future work, we would like to see more interesting work exploring new multi-task learning approaches.

\section{Conclusion}
To conclude, in this paper, we present \baby, a jointly labeled large-scale dataset for classification, question answering and summarization. We benchmark existing methods and a straightforward baseline with a novel multi-task paradigm on \baby and analyze their performance on these three tasks. Our extensive experiments reveal the potential of the proposed dataset for accelerating the innovations in the three tasks and multi-task learning.

\section*{Acknowledgments}
We are grateful for the insightful comments from the anonymous reviewers. This research was supported by National Natural Science Foundation
of China (No. 61872278). Chenliang Li is the corresponding author.

\bibliography{matinf}

\begin{thebibliography}{66}
\expandafter\ifx\csname natexlab\endcsname\relax\def\natexlab#1{#1}\fi

\bibitem[{Cao et~al.(2018)Cao, Lu, Zhou, and Li}]{cw2vec}
Shaosheng Cao, Wei Lu, Jun Zhou, and Xiaolong Li. 2018.
\newblock \href
  {https://www.aaai.org/ocs/index.php/AAAI/AAAI18/paper/view/17444} {cw2vec:
  Learning chinese word embeddings with stroke n-gram information}.
\newblock In \emph{{AAAI}}.

\bibitem[{Chen et~al.(2017)Chen, Fisch, Weston, and Bordes}]{chen2017reading}
Danqi Chen, Adam Fisch, Jason Weston, and Antoine Bordes. 2017.
\newblock \href {https://doi.org/10.18653/v1/P17-1171} {Reading wikipedia to
  answer open-domain questions}.
\newblock In \emph{{ACL}}.

\bibitem[{Cohan et~al.(2018)Cohan, Dernoncourt, Kim, Bui, Kim, Chang, and
  Goharian}]{arxiv}
Arman Cohan, Franck Dernoncourt, Doo~Soon Kim, Trung Bui, Seokhwan Kim, Walter
  Chang, and Nazli Goharian. 2018.
\newblock \href {https://doi.org/10.18653/v1/n18-2097} {A discourse-aware
  attention model for abstractive summarization of long documents}.
\newblock In \emph{{NAACL-HLT}}.

\bibitem[{Cui et~al.(2016)Cui, Liu, Chen, Wang, and Hu}]{hlfrc}
Yiming Cui, Ting Liu, Zhipeng Chen, Shijin Wang, and Guoping Hu. 2016.
\newblock \href {https://www.aclweb.org/anthology/C16-1167/} {Consensus
  attention-based neural networks for chinese reading comprehension}.
\newblock In \emph{{COLING}}.

\bibitem[{Deng et~al.(2009)Deng, Dong, Socher, Li, Li, and Li}]{imagenet}
Jia Deng, Wei Dong, Richard Socher, Li{-}Jia Li, Kai Li, and Fei{-}Fei Li.
  2009.
\newblock \href {https://doi.org/10.1109/CVPR.2009.5206848} {Imagenet: {A}
  large-scale hierarchical image database}.
\newblock In \emph{{CVPR}}.

\bibitem[{Devlin et~al.(2019)Devlin, Chang, Lee, and Toutanova}]{bert}
Jacob Devlin, Ming{-}Wei Chang, Kenton Lee, and Kristina Toutanova. 2019.
\newblock \href {https://doi.org/10.18653/v1/n19-1423} {{BERT:} pre-training of
  deep bidirectional transformers for language understanding}.
\newblock In \emph{{NAACL-HLT}}.

\bibitem[{Dong et~al.(2019)Dong, Yang, Wang, Wei, Liu, Wang, Gao, Zhou, and
  Hon}]{unilm}
Li~Dong, Nan Yang, Wenhui Wang, Furu Wei, Xiaodong Liu, Yu~Wang, Jianfeng Gao,
  Ming Zhou, and Hsiao-Wuen Hon. 2019.
\newblock \href
  {http://papers.nips.cc/paper/9464-unified-language-model-pre-training-for-natural-language-understanding-and-generation}
  {Unified language model pre-training for natural language understanding and
  generation}.
\newblock In \emph{{NeurIPS}}.

\bibitem[{Duan and Tang(2017)}]{nlpcc17:duan}
Nan Duan and Duyu Tang. 2017.
\newblock \href {https://doi.org/10.1007/978-3-319-73618-1\_86} {Overview of
  the {NLPCC} 2017 shared task: Open domain chinese question answering}.
\newblock In \emph{{NLPCC}}.

\bibitem[{Dunn et~al.(2017)Dunn, Sagun, Higgins, G{\"{u}}ney, Cirik, and
  Cho}]{searchqa}
Matthew Dunn, Levent Sagun, Mike Higgins, V.~Ugur G{\"{u}}ney, Volkan Cirik,
  and Kyunghyun Cho. 2017.
\newblock \href {http://arxiv.org/abs/1704.05179} {Searchqa: {A} new q{\&}a
  dataset augmented with context from a search engine}.
\newblock \emph{CoRR}, abs/1704.05179.

\bibitem[{Erkan and Radev(2004)}]{lexrank}
G{\"{u}}nes Erkan and Dragomir~R. Radev. 2004.
\newblock \href {https://doi.org/10.1613/jair.1523} {Lexrank: Graph-based
  lexical centrality as salience in text summarization}.
\newblock \emph{J. Artif. Intell. Res.}

\bibitem[{Grave et~al.(2017)Grave, Mikolov, Joulin, and Bojanowski}]{fasttext}
Edouard Grave, Tomas Mikolov, Armand Joulin, and Piotr Bojanowski. 2017.
\newblock \href {https://doi.org/10.18653/v1/e17-2068} {Bag of tricks for
  efficient text classification}.
\newblock In \emph{{EACL}}.

\bibitem[{Grusky et~al.(2018)Grusky, Naaman, and Artzi}]{newsroom}
Max Grusky, Mor Naaman, and Yoav Artzi. 2018.
\newblock \href {https://doi.org/10.18653/v1/n18-1065} {Newsroom: {A} dataset
  of 1.3 million summaries with diverse extractive strategies}.
\newblock In \emph{{NAACL-HLT}}.

\bibitem[{Guo et~al.(2018)Guo, Pasunuru, and Bansal}]{acl18:guo}
Han Guo, Ramakanth Pasunuru, and Mohit Bansal. 2018.
\newblock \href {https://www.aclweb.org/anthology/P18-1064/} {Soft
  layer-specific multi-task summarization with entailment and question
  generation}.
\newblock In \emph{{ACL}}.

\bibitem[{He et~al.(2018)He, Liu, Liu, Lyu, Zhao, Xiao, Liu, Wang, Wu, She,
  Liu, Wu, and Wang}]{dureader}
Wei He, Kai Liu, Jing Liu, Yajuan Lyu, Shiqi Zhao, Xinyan Xiao, Yuan Liu,
  Yizhong Wang, Hua Wu, Qiaoqiao She, Xuan Liu, Tian Wu, and Haifeng Wang.
  2018.
\newblock \href {https://www.aclweb.org/anthology/W18-2605/} {Dureader: a
  chinese machine reading comprehension dataset from real-world applications}.
\newblock In \emph{QA@ACL}.

\bibitem[{Hermann et~al.(2015)Hermann, Kocisk{\'{y}}, Grefenstette, Espeholt,
  Kay, Suleyman, and Blunsom}]{cnndm}
Karl~Moritz Hermann, Tom{\'{a}}s Kocisk{\'{y}}, Edward Grefenstette, Lasse
  Espeholt, Will Kay, Mustafa Suleyman, and Phil Blunsom. 2015.
\newblock \href
  {http://papers.nips.cc/paper/5945-teaching-machines-to-read-and-comprehend}
  {Teaching machines to read and comprehend}.
\newblock In \emph{{NeurIPS}}.

\bibitem[{Hill et~al.(2016)Hill, Bordes, Chopra, and Weston}]{cbt}
Felix Hill, Antoine Bordes, Sumit Chopra, and Jason Weston. 2016.
\newblock \href {http://arxiv.org/abs/1511.02301} {The goldilocks principle:
  Reading children's books with explicit memory representations}.
\newblock In \emph{{ICLR}}.

\bibitem[{Hu et~al.(2015)Hu, Chen, and Zhu}]{lcsts}
Baotian Hu, Qingcai Chen, and Fangze Zhu. 2015.
\newblock \href {https://doi.org/10.18653/v1/d15-1229} {{LCSTS:} {A} large
  scale chinese short text summarization dataset}.
\newblock In \emph{{EMNLP}}.

\bibitem[{Johnson and Zhang(2017)}]{dpcnn}
Rie Johnson and Tong Zhang. 2017.
\newblock \href {https://doi.org/10.18653/v1/P17-1052} {Deep pyramid
  convolutional neural networks for text categorization}.
\newblock In \emph{{ACL}}.

\bibitem[{Jurafsky and Martin(2009)}]{jurafsky2008speech}
Dan Jurafsky and James~H. Martin. 2009.
\newblock \href {http://www.worldcat.org/oclc/315913020} {\emph{Speech and
  language processing: an introduction to natural language processing,
  computational linguistics, and speech recognition, 2nd Edition}}.
\newblock Prentice Hall series in artificial intelligence. Prentice Hall,
  Pearson Education International.

\bibitem[{Kalchbrenner et~al.(2014)Kalchbrenner, Grefenstette, and
  Blunsom}]{dcnn}
Nal Kalchbrenner, Edward Grefenstette, and Phil Blunsom. 2014.
\newblock \href {https://doi.org/10.3115/v1/p14-1062} {A convolutional neural
  network for modelling sentences}.
\newblock In \emph{{ACL}}.

\bibitem[{Kim(2014)}]{textcnn}
Yoon Kim. 2014.
\newblock \href {https://doi.org/10.3115/v1/d14-1181} {Convolutional neural
  networks for sentence classification}.
\newblock In \emph{{EMNLP}}.

\bibitem[{Kingma and Ba(2015)}]{adam}
Diederik~P. Kingma and Jimmy Ba. 2015.
\newblock \href {http://arxiv.org/abs/1412.6980} {Adam: {A} method for
  stochastic optimization}.
\newblock In \emph{{ICLR}}.

\bibitem[{Lai et~al.(2015)Lai, Xu, Liu, and Zhao}]{rcnn}
Siwei Lai, Liheng Xu, Kang Liu, and Jun Zhao. 2015.
\newblock \href {http://www.aaai.org/ocs/index.php/AAAI/AAAI15/paper/view/9745}
  {Recurrent convolutional neural networks for text classification}.
\newblock In \emph{{AAAI}}.

\bibitem[{Lei et~al.(2020)Lei, He, Miao, Wu, Hong, Kan, and Chua}]{xiangnan}
Wenqiang Lei, Xiangnan He, Yisong Miao, Qingyun Wu, Richang Hong, Min{-}Yen
  Kan, and Tat{-}Seng Chua. 2020.
\newblock \href {https://doi.org/10.1145/3336191.3371769}
  {Estimation-action-reflection: Towards deep interaction between
  conversational and recommender systems}.
\newblock In \emph{{WSDM}}.

\bibitem[{Li et~al.(2019)Li, Meng, Sun, Han, Yuan, and Li}]{wordseg}
Xiaoya Li, Yuxian Meng, Xiaofei Sun, Qinghong Han, Arianna Yuan, and Jiwei Li.
  2019.
\newblock \href {https://doi.org/10.18653/v1/p19-1314} {Is word segmentation
  necessary for deep learning of chinese representations?}
\newblock In \emph{{ACL}}.

\bibitem[{Lin and Hovy(2003)}]{rouge}
Chin{-}Yew Lin and Eduard~H. Hovy. 2003.
\newblock \href {https://www.aclweb.org/anthology/N03-1020/} {Automatic
  evaluation of summaries using n-gram co-occurrence statistics}.
\newblock In \emph{{HLT-NAACL}}.

\bibitem[{Lin et~al.(2018)Lin, Sun, Ma, and Su}]{globalencoding}
Junyang Lin, Xu~Sun, Shuming Ma, and Qi~Su. 2018.
\newblock \href {https://www.aclweb.org/anthology/P18-2027/} {Global encoding
  for abstractive summarization}.
\newblock In \emph{{ACL}}.

\bibitem[{Liu et~al.(2017)Liu, Qiu, and Huang}]{adversial}
Pengfei Liu, Xipeng Qiu, and Xuanjing Huang. 2017.
\newblock \href {https://doi.org/10.18653/v1/P17-1001} {Adversarial multi-task
  learning for text classification}.
\newblock In \emph{{ACL}}.

\bibitem[{Liu et~al.(2015)Liu, Gao, He, Deng, Duh, and Wang}]{naacl15:liu}
Xiaodong Liu, Jianfeng Gao, Xiaodong He, Li~Deng, Kevin Duh, and Ye{-}Yi Wang.
  2015.
\newblock \href {https://doi.org/10.3115/v1/n15-1092} {Representation learning
  using multi-task deep neural networks for semantic classification and
  information retrieval}.
\newblock In \emph{{NAACL-HLT}}.

\bibitem[{Liu et~al.(2019)Liu, He, Chen, and Gao}]{mtdnn}
Xiaodong Liu, Pengcheng He, Weizhu Chen, and Jianfeng Gao. 2019.
\newblock \href {https://doi.org/10.18653/v1/p19-1441} {Multi-task deep neural
  networks for natural language understanding}.
\newblock In \emph{{ACL}}.

\bibitem[{Liu et~al.(2018)Liu, Chen, Deng, Zeng, Chen, Li, and Tang}]{lcqmc}
Xin Liu, Qingcai Chen, Chong Deng, Huajun Zeng, Jing Chen, Dongfang Li, and
  Buzhou Tang. 2018.
\newblock \href {https://www.aclweb.org/anthology/C18-1166/} {{LCQMC:} {A}
  large-scale chinese question matching corpus}.
\newblock In \emph{{COLING}}.

\bibitem[{Liu and Lapata(2019)}]{bertabs}
Yang Liu and Mirella Lapata. 2019.
\newblock \href {https://doi.org/10.18653/v1/D19-1387} {Text summarization with
  pretrained encoders}.
\newblock In \emph{{EMNLP/IJCNLP}}.

\bibitem[{Luong et~al.(2016)Luong, Le, Sutskever, Vinyals, and
  Kaiser}]{iclr16:luong}
Minh{-}Thang Luong, Quoc~V. Le, Ilya Sutskever, Oriol Vinyals, and Lukasz
  Kaiser. 2016.
\newblock \href {http://arxiv.org/abs/1511.06114} {Multi-task sequence to
  sequence learning}.
\newblock In \emph{{ICLR}}.

\bibitem[{Luong et~al.(2015)Luong, Pham, and Manning}]{s2satt}
Thang Luong, Hieu Pham, and Christopher~D. Manning. 2015.
\newblock \href {https://doi.org/10.18653/v1/d15-1166} {Effective approaches to
  attention-based neural machine translation}.
\newblock In \emph{{EMNLP}}.

\bibitem[{Ma et~al.(2018)Ma, Sun, Li, Li, Li, and Ren}]{wean}
Shuming Ma, Xu~Sun, Wei Li, Sujian Li, Wenjie Li, and Xuancheng Ren. 2018.
\newblock \href {https://doi.org/10.18653/v1/n18-1018} {Query and output:
  Generating words by querying distributed word representations for paraphrase
  generation}.
\newblock In \emph{{NAACL-HLT}}.

\bibitem[{McCann et~al.(2018)McCann, Keskar, Xiong, and Socher}]{decathlon}
Bryan McCann, Nitish~Shirish Keskar, Caiming Xiong, and Richard Socher. 2018.
\newblock \href {http://arxiv.org/abs/1806.08730} {The natural language
  decathlon: Multitask learning as question answering}.
\newblock \emph{CoRR}, abs/1806.08730.

\bibitem[{Mihalcea and Tarau(2004)}]{textrank}
Rada Mihalcea and Paul Tarau. 2004.
\newblock \href {https://www.aclweb.org/anthology/W04-3252/} {Textrank:
  Bringing order into text}.
\newblock In \emph{{EMNLP}}.

\bibitem[{Napoles et~al.(2012)Napoles, Gormley, and Durme}]{gigaword}
Courtney Napoles, Matthew~R. Gormley, and Benjamin~Van Durme. 2012.
\newblock \href {https://www.aclweb.org/anthology/W12-3018/} {Annotated
  gigaword}.
\newblock In \emph{AKBC-WEKEX@NAACL-HLT}.

\bibitem[{Narayan et~al.(2018)Narayan, Cohen, and Lapata}]{xsum}
Shashi Narayan, Shay~B. Cohen, and Mirella Lapata. 2018.
\newblock \href {https://doi.org/10.18653/v1/d18-1206} {Don't give me the
  details, just the summary! topic-aware convolutional neural networks for
  extreme summarization}.
\newblock In \emph{{EMNLP}}.

\bibitem[{Nguyen et~al.(2016)Nguyen, Rosenberg, Song, Gao, Tiwary, Majumder,
  and Deng}]{msmarco}
Tri Nguyen, Mir Rosenberg, Xia Song, Jianfeng Gao, Saurabh Tiwary, Rangan
  Majumder, and Li~Deng. 2016.
\newblock \href {http://ceur-ws.org/Vol-1773/CoCoNIPS\_2016\_paper9.pdf} {{MS}
  {MARCO:} {A} human generated machine reading comprehension dataset}.
\newblock In \emph{CoCo@NeurIPS}.

\bibitem[{Radford et~al.(2019)Radford, Wu, Child, Luan, Amodei, and
  Sutskever}]{gpt2}
Alec Radford, Jeffrey Wu, Rewon Child, David Luan, Dario Amodei, and Ilya
  Sutskever. 2019.
\newblock \href
  {https://d4mucfpksywv.cloudfront.net/better-language-models/language_models_are_unsupervised_multitask_learners.pdf}
  {Language models are unsupervised multitask learners}.

\bibitem[{Raffel et~al.(2019)Raffel, Shazeer, Roberts, Lee, Narang, Matena,
  Zhou, Li, and Liu}]{t5}
Colin Raffel, Noam Shazeer, Adam Roberts, Katherine Lee, Sharan Narang, Michael
  Matena, Yanqi Zhou, Wei Li, and Peter~J. Liu. 2019.
\newblock \href {http://arxiv.org/abs/1910.10683} {Exploring the limits of
  transfer learning with a unified text-to-text transformer}.
\newblock \emph{CoRR}, abs/1910.10683.

\bibitem[{Rajpurkar et~al.(2016)Rajpurkar, Zhang, Lopyrev, and Liang}]{squad}
Pranav Rajpurkar, Jian Zhang, Konstantin Lopyrev, and Percy Liang. 2016.
\newblock \href {https://doi.org/10.18653/v1/d16-1264} {Squad: 100, 000+
  questions for machine comprehension of text}.
\newblock In \emph{{EMNLP}}.

\bibitem[{Ruder et~al.(2019)Ruder, Bingel, Augenstein, and
  S{\o}gaard}]{aaai19:ruder}
Sebastian Ruder, Joachim Bingel, Isabelle Augenstein, and Anders S{\o}gaard.
  2019.
\newblock \href {https://www.aclweb.org/anthology/P18-1064/} {Latent multi-task
  architecture learning}.
\newblock In \emph{{AAAI}}.

\bibitem[{Sharma et~al.(2019)Sharma, Li, and Wang}]{bigpatent}
Eva Sharma, Chen Li, and Lu~Wang. 2019.
\newblock \href {https://doi.org/10.18653/v1/p19-1212} {{BIGPATENT:} {A}
  large-scale dataset for abstractive and coherent summarization}.
\newblock In \emph{{ACL}}.

\bibitem[{Shen et~al.(2019)Shen, Geng, Qin, Guo, Tang, Duan, Long, and
  Jiang}]{guoday}
Tao Shen, Xiubo Geng, Tao Qin, Daya Guo, Duyu Tang, Nan Duan, Guodong Long, and
  Daxin Jiang. 2019.
\newblock \href {https://doi.org/10.18653/v1/D19-1248} {Multi-task learning for
  conversational question answering over a large-scale knowledge base}.
\newblock In \emph{{EMNLP/IJCNLP}}.

\bibitem[{Song et~al.(2018)Song, Shi, Li, and Zhang}]{tencentemb}
Yan Song, Shuming Shi, Jing Li, and Haisong Zhang. 2018.
\newblock \href {https://doi.org/10.18653/v1/n18-2028} {Directional skip-gram:
  Explicitly distinguishing left and right context for word embeddings}.
\newblock In \emph{{NAACL-HLT}}.

\bibitem[{Sun et~al.(2018)Sun, Yang, Dong, Zhang, Dong, and Young}]{superchar}
Baohua Sun, Lin Yang, Patrick Dong, Wenhan Zhang, Jason Dong, and Charles
  Young. 2018.
\newblock \href {https://doi.org/10.18653/v1/w18-6245} {Super characters: {A}
  conversion from sentiment classification to image classification}.
\newblock In \emph{WASSA@EMNLP}.

\bibitem[{Sun et~al.(2016)Sun, Li, Guo, Zhao, Zheng, Si, and Liu}]{thucnews}
Maosong Sun, Jingyang Li, Zhipeng Guo, Yu~Zhao, Yabin Zheng, Xiance Si, and
  Zhiyuan Liu. 2016.
\newblock \href {http://thuctc.thunlp.org} {{THUCTC}: An efficient chinese text
  classifier}.

\bibitem[{Sun et~al.(2019)Sun, Wang, Li, Feng, Chen, Zhang, Tian, Zhu, Tian,
  and Wu}]{ernie}
Yu~Sun, Shuohuan Wang, Yu{-}Kun Li, Shikun Feng, Xuyi Chen, Han Zhang, Xin
  Tian, Danxiang Zhu, Hao Tian, and Hua Wu. 2019.
\newblock \href {http://arxiv.org/abs/1904.09223} {{ERNIE:} enhanced
  representation through knowledge integration}.
\newblock \emph{CoRR}, abs/1904.09223.

\bibitem[{Sutskever et~al.(2014)Sutskever, Vinyals, and Le}]{s2s}
Ilya Sutskever, Oriol Vinyals, and Quoc~V. Le. 2014.
\newblock \href
  {http://papers.nips.cc/paper/5346-sequence-to-sequence-learning-with-neural-networks}
  {Sequence to sequence learning with neural networks}.
\newblock In \emph{{NeurIPS}}.

\bibitem[{Trischler et~al.(2017)Trischler, Wang, Yuan, Harris, Sordoni,
  Bachman, and Suleman}]{newsqa}
Adam Trischler, Tong Wang, Xingdi Yuan, Justin Harris, Alessandro Sordoni,
  Philip Bachman, and Kaheer Suleman. 2017.
\newblock \href {https://doi.org/10.18653/v1/w17-2623} {Newsqa: {A} machine
  comprehension dataset}.
\newblock In \emph{Rep4NLP@ACL}.

\bibitem[{Voorhees and Tice(1999)}]{trec}
Ellen~M. Voorhees and Dawn~M. Tice. 1999.
\newblock \href {http://trec.nist.gov/pubs/trec8/papers/qa8.pdf} {The {TREC-8}
  question answering track evaluation}.
\newblock In \emph{{TREC}}.

\bibitem[{Wang et~al.(2019{\natexlab{a}})Wang, Pruksachatkun, Nangia, Singh,
  Michael, Hill, Levy, and Bowman}]{superglue}
Alex Wang, Yada Pruksachatkun, Nikita Nangia, Amanpreet Singh, Julian Michael,
  Felix Hill, Omer Levy, and Samuel~R. Bowman. 2019{\natexlab{a}}.
\newblock \href
  {http://papers.nips.cc/paper/8589-superglue-a-stickier-benchmark-for-general-purpose-language-understanding-systems}
  {Superglue: {A} stickier benchmark for general-purpose language understanding
  systems}.
\newblock In \emph{NeurIPS}.

\bibitem[{Wang et~al.(2019{\natexlab{b}})Wang, Singh, Michael, Hill, Levy, and
  Bowman}]{glue}
Alex Wang, Amanpreet Singh, Julian Michael, Felix Hill, Omer Levy, and
  Samuel~R. Bowman. 2019{\natexlab{b}}.
\newblock \href {https://openreview.net/forum?id=rJ4km2R5t7} {{GLUE:} {A}
  multi-task benchmark and analysis platform for natural language
  understanding}.
\newblock In \emph{{ICLR}}.

\bibitem[{Wang et~al.(2008{\natexlab{a}})Wang, Zhang, Ma, and Ru}]{sogou}
Canhui Wang, Min Zhang, Shaoping Ma, and Liyun Ru. 2008{\natexlab{a}}.
\newblock \href {https://doi.org/10.1145/1367497.1367560} {Automatic online
  news issue construction in web environment}.
\newblock In \emph{{WWW}}.

\bibitem[{Wang et~al.(2008{\natexlab{b}})Wang, Zhang, Ma, and Ru}]{sogoucs}
Canhui Wang, Min Zhang, Shaoping Ma, and Liyun Ru. 2008{\natexlab{b}}.
\newblock \href {https://doi.org/10.1145/1367497.1367560} {Automatic online
  news issue construction in web environment}.
\newblock In \emph{{WWW}}.

\bibitem[{Wiseman and Rush(2016)}]{beamsearch}
Sam Wiseman and Alexander~M. Rush. 2016.
\newblock \href {https://doi.org/10.18653/v1/d16-1137} {Sequence-to-sequence
  learning as beam-search optimization}.
\newblock In \emph{{EMNLP}}.

\bibitem[{Wu et~al.(2019)Wu, Meng, Han, Li, Li, Mei, Nie, Sun, and Li}]{glyce}
Wei Wu, Yuxian Meng, Qinghong Han, Muyu Li, Xiaoya Li, Jie Mei, Ping Nie,
  Xiaofei Sun, and Jiwei Li. 2019.
\newblock \href
  {http://papers.nips.cc/paper/8542-glyce-glyph-vectors-for-chinese-character-representations}
  {Glyce: Glyph-vectors for chinese character representations}.
\newblock In \emph{NeurIPS}.

\bibitem[{Xu et~al.(2020)Xu, Zhou, Ge, Wei, and Zhou}]{theseus}
Canwen Xu, Wangchunshu Zhou, Tao Ge, Furu Wei, and Ming Zhou. 2020.
\newblock \href {http://arxiv.org/abs/2002.02925} {Bert-of-theseus: Compressing
  {BERT} by progressive module replacing}.
\newblock \emph{CoRR}, abs/2002.02925.

\bibitem[{Xu et~al.(2019)Xu, Liu, Shen, Liu, and Gao}]{naacl19:xu}
Yichong Xu, Xiaodong Liu, Yelong Shen, Jingjing Liu, and Jianfeng Gao. 2019.
\newblock \href {https://doi.org/10.18653/v1/n19-1271} {Multi-task learning
  with sample re-weighting for machine reading comprehension}.
\newblock In \emph{{NAACL-HLT}}.

\bibitem[{Yang et~al.(2019)Yang, Dai, Yang, Carbonell, Salakhutdinov, and
  Le}]{xlnet}
Zhilin Yang, Zihang Dai, Yiming Yang, Jaime~G. Carbonell, Ruslan Salakhutdinov,
  and Quoc~V. Le. 2019.
\newblock \href
  {http://papers.nips.cc/paper/8812-xlnet-generalized-autoregressive-pretraining-for-language-understanding}
  {Xlnet: Generalized autoregressive pretraining for language understanding}.
\newblock In \emph{NeurIPS}.

\bibitem[{Zhang et~al.(2019)Zhang, Bai, Liang, Bai, Chang, Yu, Zhu, and
  Zhao}]{leakage}
Guanhua Zhang, Bing Bai, Jian Liang, Kun Bai, Shiyu Chang, Mo~Yu, Conghui Zhu,
  and Tiejun Zhao. 2019.
\newblock \href {https://doi.org/10.18653/v1/p19-1435} {Selection bias
  explorations and debias methods for natural language sentence matching
  datasets}.
\newblock In \emph{{ACL}}.

\bibitem[{Zhang and LeCun(2017)}]{ifengNchinanews}
Xiang Zhang and Yann LeCun. 2017.
\newblock \href {http://arxiv.org/abs/1708.02657} {Which encoding is the best
  for text classification in chinese, english, japanese and korean?}
\newblock \emph{CoRR}, abs/1708.02657.

\bibitem[{Zhang et~al.(2015)Zhang, Zhao, and LeCun}]{agnewsNdbpediaNyahoo}
Xiang Zhang, Junbo~Jake Zhao, and Yann LeCun. 2015.
\newblock \href
  {http://papers.nips.cc/paper/5782-character-level-convolutional-networks-for-text-classification}
  {Character-level convolutional networks for text classification}.
\newblock In \emph{{NeurIPS}}.

\bibitem[{Zhang and Zhao(2018)}]{gaokao}
Zhuosheng Zhang and Hai Zhao. 2018.
\newblock \href {https://www.aclweb.org/anthology/C18-1038/} {One-shot learning
  for question-answering in gaokao history challenge}.
\newblock In \emph{{COLING}}.

\end{thebibliography}
\bibliographystyle{acl_natbib}

\end{document}